\documentclass{article}
\usepackage[dvipsnames,table]{xcolor}
\usepackage[preprint]{corl_2026} % Use this for the initial submission.
\usepackage{microtype}
\usepackage{graphicx}
\usepackage{subcaption}
\usepackage{booktabs}
\usepackage{hyperref}
\usepackage{float}
\usepackage{wrapfig}
\usepackage{lipsum}
\usepackage{colortbl}
\usepackage{soul}
\usepackage{enumitem}
\usepackage{multirow}
\usepackage{makecell}
\usepackage{tablefootnote}
\usepackage{amsmath}
\usepackage{amssymb}
\usepackage{mathtools}
\usepackage{amsthm}
\usepackage{adjustbox}
\usepackage[capitalize,noabbrev]{cleveref}

\usepackage{amsmath,amsfonts,bm}

\def\eqref#1{equation~\ref{#1}}
\def\1{\bm{1}}

\def\rva{{\mathbf{a}}}

\def\rvh{{\mathbf{h}}}

\def\rvm{{\mathbf{m}}}

\def\rvs{{\mathbf{s}}}

\def\rmA{{\mathbf{A}}}

\def\rmI{{\mathbf{I}}}

\def\rmM{{\mathbf{M}}}

\def\vzero{{\bm{0}}}

\def\mK{{\bm{K}}}

\def\mQ{{\bm{Q}}}

\def\mV{{\bm{V}}}

\DeclareMathAlphabet{\mathsfit}{\encodingdefault}{\sfdefault}{m}{sl}
\SetMathAlphabet{\mathsfit}{bold}{\encodingdefault}{\sfdefault}{bx}{n}

\newcommand{\E}{\mathbb{E}}

\newcommand{\R}{\mathbb{R}}

\theoremstyle{plain}

\theoremstyle{definition}

\theoremstyle{remark}

\usepackage[textsize=tiny]{todonotes}

\definecolor{cornellred}{rgb}{0.7, 0.11, 0.11}
\definecolor{cadmiumgreen}{rgb}{0.0, 0.42, 0.24}
\definecolor{aliceblue}{rgb}{0.91, 0.94, 0.97}
\definecolor{darkblue}{rgb}{0.83, 0.89, 0.97}
\definecolor{Red7}{rgb}{0.941, 0.243, 0.243}
\definecolor{Green7}{rgb}{0.40, 0.710, 0.302}
\definecolor{Blue9}{rgb}{0.098, 0.3, 0.9}

\usepackage{pifont}
\newcommand{\cmark}{\ding{51}}
\newcommand{\xmark}{\ding{55}}
\newcommand{\ck}{\color{Green7}{\cmark}}
\newcommand{\xk}{\color{Red7}{\xmark}}

\newcommand{\ie}{\textit{i}.\textit{e}., }
\newcommand{\eg}{\textit{e}.\textit{g}., }

\title{Modular Sensory Stream for Integrating Physical Feedback in Vision-Language-Action Models}

\newcommand{\lname}{\textbf{Mo}dular \textbf{S}ensory \textbf{S}tream  framework}
\newcommand{\sname}{MoSS}

\author{
  Jimin Lee$^{1}$, Huiwon Jang$^{1,2}$, Myungkyu Koo$^{1,2}$, Jungwoo Park$^{1}$, Jinwoo Shin$^{1,2}$ \\[3pt]
  $^{1}$KAIST \quad $^{2}$RLWRLD \\
  \texttt{\{jiminl, jinwoos\}@kaist.ac.kr}
}

\begin{document}
\maketitle

%===============================================================================

\begin{abstract}
Humans understand and interact with the real world by relying on diverse physical feedback beyond visual perception.
Motivated by this, recent approaches attempt to incorporate physical sensory signals into Vision-Language-Action models (VLAs).
However, they typically focus on a single type of physical signal, failing to capture the heterogeneous and complementary nature of real-world interactions.
In this paper, we propose \sname, a modular sensory stream framework that adapts VLAs to leverage multiple sensory signals for action prediction.
Specifically, we introduce decoupled modality streams that integrate heterogeneous physical signals into the action stream via joint cross-modal self-attention.
To enable stable incorporation of new modalities, we adopt a two-stage training scheme that freezes pretrained VLA parameters in the early stage.
Furthermore, to better capture contact interaction dynamics, we incorporate an auxiliary task that predicts future physical signals.
Through extensive real-world experiments, we demonstrate that \sname{} successfully augments VLAs to leverage diverse physical signals (\ie tactile, force, and torque), integrating multiple signals to achieve synergistic performance gains.
The \href{https://jiminlx.github.io/MoSS/}{project page} is available.

\end{abstract}
\section{Introduction}

%%%%% 1. SHORT INTRO
In the real world, humans perform dexterous manipulation by continuously integrating multiple sources of physical sensory feedback beyond visual perception.
For example, during plug insertion, tactile feedback at the fingertips helps recognize grasp stability, while torque feedback from the arm provides crucial cues for detecting contact and correcting misalignment.
Toward learning human-like generalizable policies, Vision-Language-Action models (VLAs) \citep{zitkovich2023rt, kim2024openvla, intelligence2025pi_, huang2025thinkact, chen2025villa0x0, nvidia2025gr00t, zheng2025x} have recently shown remarkable progress by leveraging large-scale pretrained Vision-Language Models (VLMs; \citep{beyer2024paligemma, bai2025qwen2, chen2025eagle}).
However, despite this progress, VLAs generate actions from visual observations alone, which fundamentally limits their ability to handle contact-rich manipulation.

\begin{figure*}[t]
    \centering\small
    \includegraphics[width=.95\linewidth]{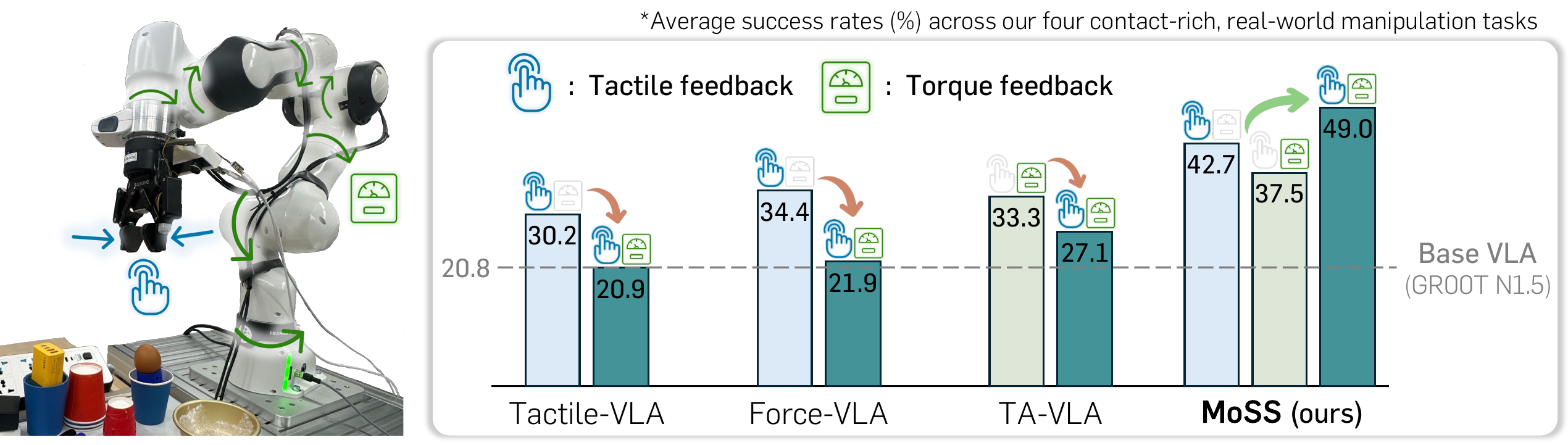}
    \caption{\textbf{Limitation of existing approaches to incorporating physical sensory signals into VLAs.}
    We compare the average success rates (\%) of VLAs as more diverse physical sensory signals are incorporated.
    Existing approaches \citep{huang2025tactile,yu2025forcevla,zhang2025ta} struggle to handle multiple physical sensing modalities, leading to degraded performance when additional modalities are used together.
    In contrast, \sname{} addresses this limitation with a unified adaptation framework that effectively integrates additional physical sensory signals into VLAs, enabling complementary use of heterogeneous physical feedback.}
    \vspace{-0.1in}
    \label{fig:teaser}
\end{figure*}
%%%%% 2. PRIOR WORK & PROBLEM
To mitigate this limitation, recent approaches have explored incorporating a single physical sensory modality into pretrained VLAs, either by encoding it into VLM backbones~\citep{zhang2025vtla, huang2025tactile, cheng2025omnivtla} or by directly conditioning the action decoder~\citep{yu2025forcevla, zhang2025ta}.
While effective in specific settings, these approaches are not designed to grow with increasing sensory complexity; 
% designed to grow: architecture explanation is needed
na\"ively extending them to incorporate additional physical modalities does not consistently yield complementary performance gains in practice (see \cref{fig:teaser}).
This highlights the need for a more unified adaptation framework that can effectively integrate heterogeneous physical signals for reliable dexterous manipulation.

%%%%% 3. METHOD
\vspace{0.01in}
{\bf Our approach.}
We present \textbf{\sname}, a \textbf{Mo}dular \textbf{S}ensory \textbf{S}tream framework that seamlessly augments pretrained VLAs to leverage diverse physical sensory signals for action prediction.
Specifically, we append decoupled modality streams to the action expert module, which processes physical inputs and exchanges information with the pretrained parameters via a joint cross-modal self-attention mechanism.
To avoid corruption of pretrained parameters by the newly added streams, we adopt a two-stage training scheme.
We first freeze the pretrained VLA and train only the physical signal streams, allowing them to pre-align their representations with the existing representation space of the pretrained policy.
We then jointly fine-tune the entire model end-to-end.
Furthermore, we incorporate an auxiliary future physical signal prediction objective, which encourages the model to internalize physical interaction dynamics and effectively utilize physical feedback for action generation \citep{zhang2025ta}.

%%%%% 4. EXPERIMENT
To validate the effectiveness of \sname, we evaluate our method on contact-rich, real-world manipulation tasks that require physical interaction with surrounding environments for precise control.
We show that \sname{} consistently improves the performance of pretrained VLAs by effectively leveraging physical feedback from both end-effector sensing (\ie tactile) and joint-level sensing (\ie torque), compared to other baselines.
Furthermore, we demonstrate that \sname{} can jointly incorporate multiple physical signals within a single model, achieving cumulative performance gains by combining complementary signals.
To the best of our knowledge, we are the first to present a unified VLA framework that simultaneously integrates multiple physical signals.

%%%%% 5. SUMMARY
We summarize the main contributions of this paper below:
\vspace{-0.1in}
\begin{itemize}[leftmargin=*,itemsep=0mm]
    \item We propose \sname, a modular framework that augments pretrained VLAs to leverage physical sensory signals via decoupled modality streams and joint cross-modal attention mechanisms.
    \item We introduce a two-stage training scheme with auxiliary future physical signal prediction objective, which preserves pretrained VLA priors while ensuring effective use of newly added modalities.
    \item We design contact-rich, real-world manipulation tasks with diverse physical feedback (\ie tactile, force, and torque) across both gripper and dexterous hand embodiments, and show that \sname{} achieves consistent and cumulative performance gains over the baselines.
\end{itemize}

\begin{figure*}[t]
    \centering\small
    \includegraphics[width=.9\linewidth]{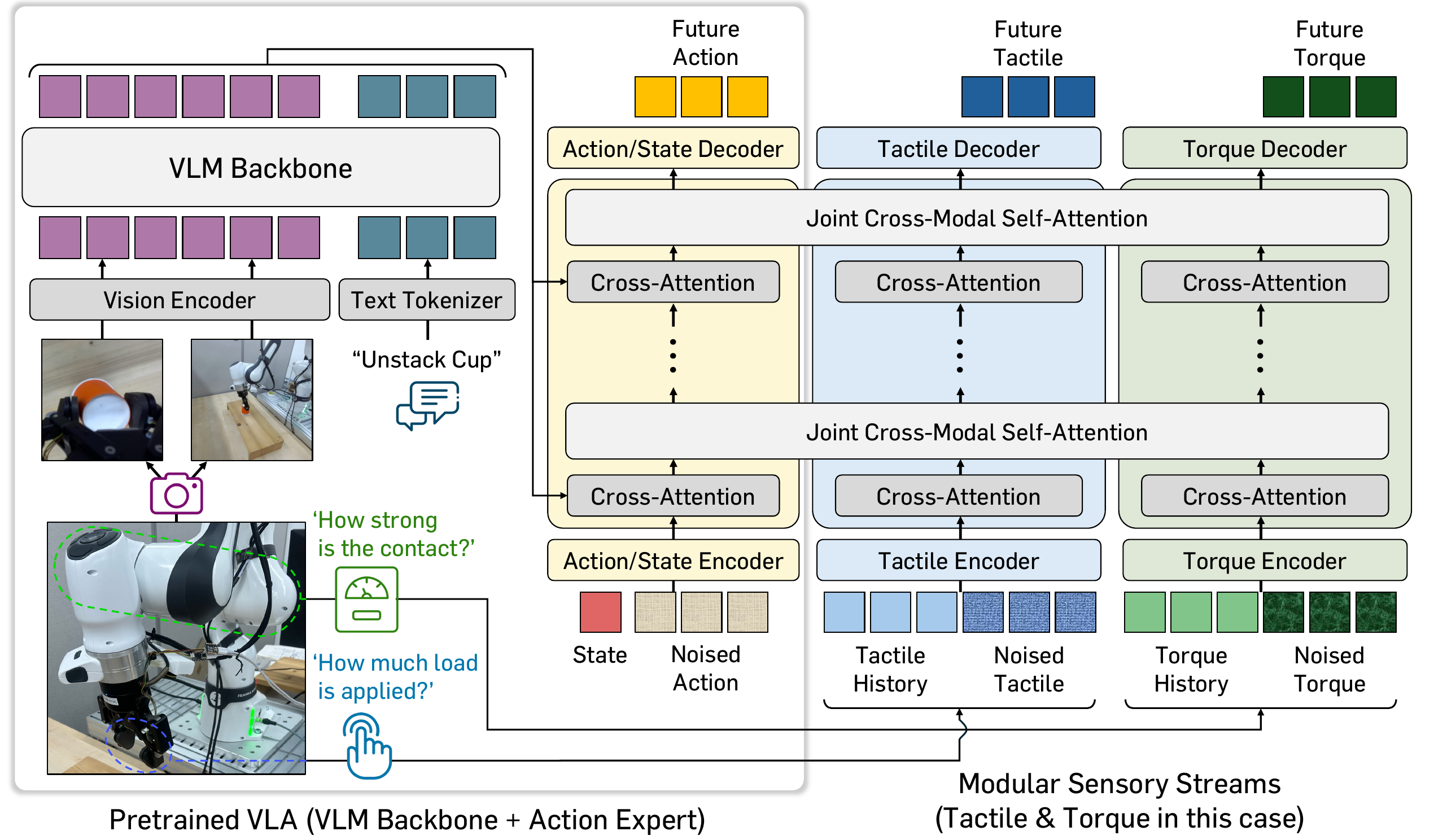}
    \vspace{-0.05cm}
    \caption{\textbf{Overview of \lname{} (\sname).} Building on a pretrained VLA, \sname{} introduces a multimodal stream architecture that processes multiple physical sensory signals (\eg tactile and torque) in parallel. We extend the self-attention layers of the Action Expert into joint cross-modal self-attention layers, enabling each signal to be processed in a dedicated stream while allowing information exchange only through joint attention.
    }
    \label{fig:overview}
    \vspace{-0.14in}
\end{figure*}
\section{Related Work}
We discuss the most relevant literature here and provide further discussion in Appendix~\ref{appendix:related-work}.

\vspace{0.01in}
{\bf Vision-Language-Action models (VLAs).}
The development of generalist robot policies has been fundamentally shaped by the challenge of acquiring diverse, large-scale robotic datasets.
While traditional approaches~\citep{shafiullah2022behavior, chi2023diffusion, lee2024behavior} rely on task-specific demonstrations with limited generalization, VLAs address this bottleneck by leveraging pretrained Vision-Language Models (VLMs) for robotic control.
Early VLAs~\citep{zitkovich2023rt, kim2024openvla} generate discretized actions autoregressively as tokens, while more recent work~\citep{intelligence2025pi_, nvidia2025gr00t} has shifted toward hybrid architectures in which diffusion-based action heads are conditioned on VLM representations.
Despite these advances, current VLAs predominantly rely on vision and language modalities, overlooking physical feedback (\eg tactile and torque) that is critical for contact-rich manipulation.
In this work, we focus on diffusion-based VLAs and introduce an approach that integrates diverse physical sensory signals into their action generation process.

\vspace{0.01in}
{\bf Vision-Language-Action models (VLAs) with physical feedback.}
Recent approaches have explored extending pretrained VLAs with physical feedback for dexterous manipulation using either \emph{end-effector-level sensing}, \eg tactile or force feedback from the gripper \citep{yu2025forcevla, huang2025tactile, bi2025vla, cheng2025omnivtla, adeniji2025feel, lin2025pp,zhang2025vtla}, or \emph{joint-level sensing}, \eg torque feedback from robot arm joints \citep{zhang2025ta}. However, existing works are specialized for a single sensing modality or sensor configurations, while dexterous manipulation often benefits from complementary physical feedback. For instance, end-effector-level sensing provides fingertip-level contact perception and has been used for tasks requiring fine contact understanding, while joint-level sensing captures whole-arm interaction cues, such as object weight, which are useful for compliant and contact-aware control.
In contrast, our framework provides a unified approach to incorporate multiple physical signals into pretrained VLAs, enabling generalist policies that can flexibly leverage available physical sensing modalities across tasks and hardware configurations.
\section{Method}

In this section, we present \sname, an adaptation framework that augments pretrained Vision-Language-Action models (VLAs) with physical sensory modalities.
Formally, let $\ell$ be a task instruction and $\rmI_t$ be the visual observations at timestep $t$.
A diffusion-based VLA processes these inputs with its Vision-Language Model (VLM) backbone $f_{\theta}$ to obtain a unified visual-language representation 
$\rvh_t = f_{\theta}(\ell, \rmI_t)\text{.}$
This feature is then passed to the action expert module $V_{\psi}$ alongside the robot's proprioceptive state $\rvs_t$ to predict $H$ future actions $\rmA_t = [\rva_t, \rva_{t+1}, \ldots, \rva_{t+H-1}]$, namely action chunking ~\citep{zhao2023learning, chi2023diffusion}.
The model is trained with a conditional flow-matching objective:
\begin{equation}
    \begin{aligned}
    \mathcal{L}=\E_{\tau, \epsilon} \left[ \| (\rmA_t - \epsilon) - V_{\psi}(\rmA_t^\tau, \rvs_t~|~\rvh_t) \|^2 \right]\text{,}
    \end{aligned}
    \label{eq:objective}
\end{equation}
where the noisy action chunk is given by $\rmA_t^\tau=\tau\rmA_t+(1-\tau)\epsilon$, with random noise $\epsilon\sim\mathcal{N}(\vzero,\rmI)$ and flow-matching timestep $\tau\in\left[0,1\right]$.

Here, our goal is to augment such VLAs to leverage a diverse set of physical feedback in addition to visual observations.
To this end, we first introduce a multimodal stream architecture that processes newly added physical signals in parallel via decoupled modality streams (see \cref{sec:method:architecture}).
We then detail our two-stage training scheme to ensure stable incorporation of these modality streams (see \cref{sec:method:2stage}), and an auxiliary future feedback prediction objective to exploit physical signals better (see \cref{sec:method:objective}).
An overview of our framework is shown in \cref{fig:overview}.

\subsection{Multimodal Sensory Streams} \label{sec:method:architecture}
Let $\rmM=\{\rvm_t^{(i)} \in \R^{d_i}\}_{i=1}^N$ be the set of $N$ physical sensory signals (\eg tactile and torque) at timestep $t$.
To incorporate them into a pretrained VLA, we extend its transformer-based action expert module $V_\psi$ with modular sensory streams $V_{\phi_i}$, each dedicated to processing the $i$-th physical signal $\rvm_t^{(i)}$.
Specifically, all sensory streams take the visual-language representation $\rvh_t$, with cross-modal interaction restricted to the self-attention layers interleaved throughout the network.
\begin{equation}
    \begin{aligned}
    &\text{Action stream}: &&\hspace{-0.6em}
    V_{\psi}(\rmA_t^\tau, \rvs_t~|~\{\rvh_t\}\cup\rmM) \\
    &\text{Physical stream } i: &&\hspace{-0.6em} 
    V_{\phi_i}(\rvm_t^{(i)}~|~\{\rvh_t, \rmA_t^\tau, \rvs_t\}\cup\rmM \backslash \{\rvm_{t}^{(i)}\})
    \end{aligned}
    \label{eq:streams}
\end{equation}
In practice, we construct each new sensory stream $V_{\phi_i}$ by mirroring the architecture of the original action expert module $V_\psi$ and randomly initializing its parameters.
We then replace the self-attention layers in each stream with joint cross-modal self-attention layers.
Specifically, at each layer, independently computed queries, keys, and values $\{\mQ_i, \mK_i, \mV_i\}_{i=0}^N$ (with $i=0$ denoting the action stream) are concatenated to calculate a shared scaled dot-product attention.
This bidirectional design enables cross-modal reasoning (\eg modulating actions based on tactile feedback or unexpected torques), while keeping other network components decoupled to prevent gradient interference.

\subsection{Two-Stage Training Strategy}\label{sec:method:2stage}
We employ a two-stage training scheme designed to preserve pretrained knowledge while enabling effective learning of physical sensory representations.

\vspace{0.01in}
{\bf Stage~1: Physical alignment.}
We freeze all parameters of the pretrained action expert module $V_\psi$ and train only the randomly initialized sensory streams $\{V_{\phi_i}\}_{i=1}^N$.
This allows the new streams to learn physical interaction features (\eg tactile contacts, force feedback, and joint torques) and cross-modal correspondences through joint self-attention, without disrupting the pretrained action policy.

\vspace{0.01in}
{\bf Stage~2: Joint fine-tuning.}
After initializing the modular sensory streams $\{V_{\phi_i}\}_{i=1}^N$, we unfreeze the action expert module $V_\psi$ and perform joint fine-tuning.
This enables the action stream to incorporate physical feedback $\rmM$ while the sensory streams adapt to task-specific requirements, jointly learning coordinated behaviors for physically-aware manipulation.

Overall, this staged strategy is crucial for stable optimization, as directly performing end-to-end training leads to optimization difficulties due to gradient interference between the randomly initialized sensory streams $\{V_{\phi_i}\}_{i=1}^N$ and pretrained action stream $V_\psi$ (see \cref{sec:exp:analysis}).

\subsection{Training Objective}\label{sec:method:objective}
Our primary objective is the action flow-matching loss
$    \mathcal{L}_{\mathrm{act}} = \E_{\tau, \epsilon} \big[ \| (\rmA_t - \epsilon) - \hat{\rmA}_t \|^2 \big]$,
where $\hat{\rmA}_t=V_{\psi}(\rmA_t^\tau, \rvs_t \mid \{\rvh_t\}\cup\rmM)$.
To encourage effective use of physical sensory signals, each modality stream predicts the future signal over the action horizon from past signals $\rvm_{t-H+1:t}^{(i)}$:
\setlength{\abovedisplayskip}{4pt}
\setlength{\belowdisplayskip}{4pt}
\setlength{\abovedisplayshortskip}{2pt}
\setlength{\belowdisplayshortskip}{2pt}
\begin{equation}
    \mathcal{L}_{\mathrm{phy}} = \sum_{i=1}^N \E_{\tau, \epsilon_i} \big[ \| (\rvm_{t+1:t+H}^{(i)} - \epsilon_i) - \hat{\rvm}_{t+1:t+H}^{(i)} \|^2 \big].
\end{equation}
The overall objective is $\mathcal{L} = \mathcal{L}_{\mathrm{act}} + \lambda_{\mathrm{phy}} \mathcal{L}_{\mathrm{phy}}$, where $\lambda_{\mathrm{phy}}$ balances the two terms. As described in \cref{sec:method:2stage}, Stage~1 optimizes only $\mathcal{L}_{\mathrm{phy}}$ with the action expert frozen, while Stage~2 jointly optimizes the full objective with respect to all parameters.
\begin{figure}[t]
    \centering\small
    \includegraphics[width=.95\linewidth]{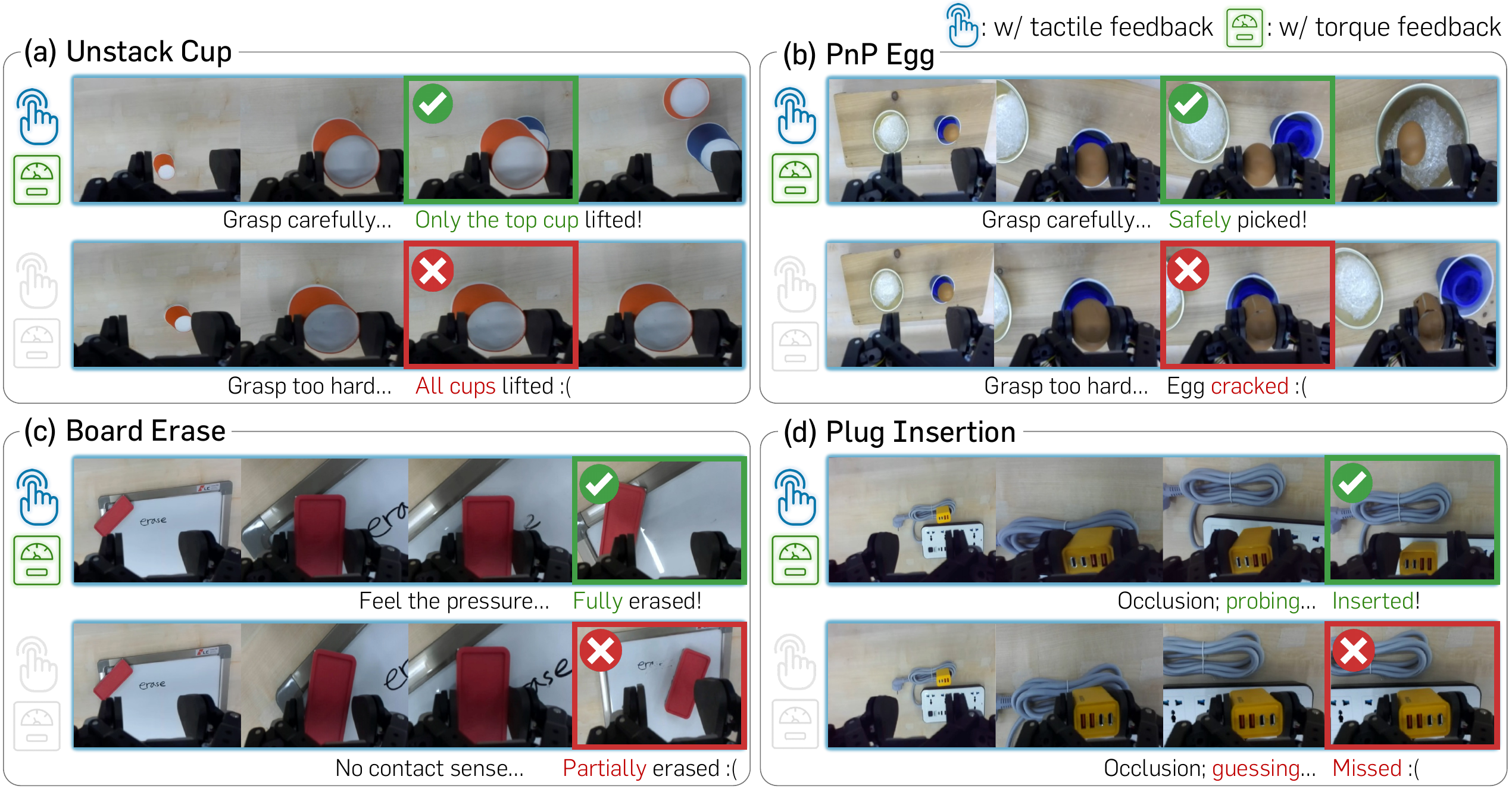}
    % \vspace{-0.2cm}
    \caption{\textbf{Example rollouts of real-world tasks.}
    We provide example rollouts of the designed tasks that critically depend on physical feedback.
    While \sname{} leverages physical feedback to successfully perform the tasks, GR00T N1.5 without physical feedback often has difficulties in (a), (b) regulating grasp force, (c) maintaining appropriate pushing force for contact, and (d) probing occluded geometry.
    }
    \label{fig:qualitative}
    \vspace{-0.05in}
\end{figure}
\section{Experiments}
We investigate the following questions through experiments:
\vspace{-0.5em}
\begin{itemize}[leftmargin=*, itemsep=0mm]
    \item Does applying \sname{} to existing Vision-Language-Action models (VLAs) effectively augment the base policy with diverse physical feedback for contact-rich manipulation, outperforming prior modality-augmented VLA baselines? (\Cref{tab:main,fig:qualitative})
    \item Can \sname{} effectively integrate multiple physical sensing signals to exploit their complementary effects? (\cref{tab:main})
    \item How do individual components affect performance? (\cref{tab:ablation,tab:efficiency,fig:attention})
    \item Does \sname{} transfer to different embodiments (\eg hand) and physical feedback (\eg force signal) with distinct sensor configurations? (\Cref{tab:hand})
\end{itemize}

\subsection{Experimental Setups}

\vspace{0.01in}
{\bf Tasks.}
We evaluate \sname{} on real-world, contact-rich manipulation tasks where interaction with physical feedback (\eg tactile on the end-effector and torque on arm joints) is crucial for successful execution.
Specifically, we design the following tasks (see \cref{fig:qualitative} for the visualization of the tasks, and see Appendix~\ref{appendix:tasks} for more detailed descriptions):
\vspace{-0.5em}
\begin{itemize}[leftmargin=*,itemsep=0mm]
    \item \textbf{Unstack Cup. } A robot should remove only the top red cup from a stack of 3 inverted cups. This task requires regulating grasping force, as gripping too strongly lifts multiple cups together.
    \item \textbf{PnP Egg. } A robot must pick up a fragile egg and place it in a bowl without breaking it. This task requires fine-grained feedback at the end-effector to avoid excessive force.
    \item \textbf{Board Erase. } A robot must erase a marked region on a whiteboard using an eraser. This task requires sensing contact to maintain consistent pressure on the board to ensure effective erasing.
    \item \textbf{Plug Insertion. } A robot must insert a plug into a socket, often occluded from the (wrist-mounted) camera view. This task requires torque and force cues to detect contact and guide its alignment.
\end{itemize}

\begin{table*}[t]
  \centering
  \caption{\textbf{Results on contact-rich, real-world manipulation tasks.}
  We report the success rates (\%) for each task using VLAs fine-tuned on a consolidated dataset that combines the training data from all tasks.
  For the averaged success rates (Avg.), we also report a standard deviation computed over all trials. 
  \textbf{Bold} and \underline{underline} indicate best and runner-up results, respectively.}
  \label{tab:main}
\vspace{-0.5em}
  \begin{adjustbox}{max width=0.95\textwidth}
  \begin{tabular}{lccccccc}
    \toprule
    Method
      & Tactile & Torque
      & Unstack Cup & PnP Egg & Board Erase
      & Plug Insertion & \textbf{Avg.} \\
    \midrule
    
    GR00T N1.5~\citep{nvidia2025gr00t}
      & \xk & \xk
      & 16.7 & 45.8 & 20.8 & \phantom{0}0.0 
      & 20.8 {\scriptsize $\pm$ 4.1}  \\
    \arrayrulecolor{black!40}\midrule
    + Tactile-VLA~\citep{huang2025tactile}
      & \ck & \xk
      & 29.2 & 45.8 &	33.3 & 12.5 
      & 30.2 {\scriptsize $\pm$ 4.7} \\
    + ForceVLA~\citep{yu2025forcevla}
      & \ck & \xk
      & 37.5 & \underline{54.2} & 33.3 & 12.5 
      & 34.4 {\scriptsize $\pm$ 4.8} \\
    \textbf{+ \sname{} (ours)}
      & \ck & \xk
      & \underline{45.8} & \textbf{66.7} & \underline{41.7} & 16.7 
      & \underline{42.7} {\scriptsize $\pm$ 5.0} \\
    \midrule
    + TA-VLA~\citep{zhang2025ta}
      & \xk & \ck
      & 29.2 & 45.8 & 37.5 & \underline{20.8} 
      & 33.3 {\scriptsize $\pm$ 4.7} \\
    \textbf{+ \sname{} (ours)}
      & \xk & \ck
      & 33.3 & 50.0 & \underline{41.7} & \textbf{25.0} 
      & 37.5 {\scriptsize $\pm$ 4.9} \\
    \midrule
    \rowcolor{gray!10}
    \textbf{+ \sname{} (ours)}
      & \ck & \ck
      & \textbf{54.2} & \textbf{66.7} & \textbf{50.0} & \textbf{25.0}
      & \textbf{49.0} {\scriptsize $\pm$ 5.1}\\    

     \arrayrulecolor{black}\midrule
    
    $\pi_{0}$~\citep{black2024pi_0}
      & \xk & \xk
      & 12.5 & 50.0 & 29.2 & 16.7 
      & 27.1 {\scriptsize $\pm$ 4.5} \\
    \arrayrulecolor{black!40}\midrule
     + Tactile-VLA~\citep{huang2025tactile}
      & \ck & \xk
      & 16.7 & 50.0 & 37.5 & 12.5 
      & 29.2 {\scriptsize $\pm$ 4.6}\\
     + ForceVLA~\citep{yu2025forcevla}
      & \ck & \xk
      & 25.0 & \underline{62.5} & 50.0 & 16.7 & 38.6 {\scriptsize $\pm$ 5.0}\\
     \textbf{+ \sname{} (ours)}
      & \ck & \xk
      & \textbf{29.2} & \underline{62.5} & 41.7 & \underline{20.8} & 38.6 {\scriptsize $\pm$ 5.0}\\
     \midrule
     + TA-VLA~\citep{zhang2025ta}
      & \xk & \ck
      & 16.7 & 58.3 & 41.7 & \underline{20.8} & 34.4 {\scriptsize $\pm$ 4.8}\\
     \textbf{+ \sname{} (ours)}
      & \xk & \ck
      & \underline{20.8} & \underline{62.5} & \underline{54.2} & \textbf{29.2} & \underline{41.7} {\scriptsize $\pm$ 5.0}\\
     \midrule
     \rowcolor{gray!10}
     \textbf{+ \sname{} (ours)}
      & \ck & \ck
      & \textbf{29.2} & \textbf{66.7} & \textbf{58.3} & \textbf{29.2} & \textbf{45.9} {\scriptsize $\pm$ 5.1} \\

    \arrayrulecolor{black}\bottomrule
  \end{tabular}
  \end{adjustbox}
  % \vspace{-0.2in}
\end{table*}
\vspace{0.01in}
{\bf Hardware setup.}
We construct a Franka Research 3 platform, a 7-DoF single arm robot equipped with a Robotiq 2F-85 gripper, following the DROID~\citep{khazatsky2024droid} setup.
Here, we mount an AnySkin~\citep{bhirangi2025anyskin} tactile sensor on the gripper, while the robot arm itself supports torque sensing at each joint.
We provide detailed specifications of the physical sensory signals in Appendix~\ref{appendix:hardware}.

\vspace{0.01in}
{\bf Baselines.}
To demonstrate the generality of our approach, we apply \sname{} to two state-of-the-art VLAs: $\pi_0$~\citep{black2024pi_0} and GR00T N1.5~\citep{nvidia2025gr00t}.
We compare \sname{} with several modality-augmented VLA baselines:
(i) Tactile-VLA~\citep{huang2025tactile} incorporates \emph{tactile} signals on the gripper as additional input tokens to the VLM backbone;
(ii) ForceVLA~\citep{yu2025forcevla} employs an MoE-based encoder to fuse 3-axis \emph{force} signals on the gripper with VLM features and conditions the action expert on them;
(iii) TA-VLA~\citep{zhang2025ta} injects \emph{torque} signals from robotic arm joints as additional tokens into a single action expert transformer shared with action tokens, while training the policy to predict future torque signals as an auxiliary task.
We adapt each method to use the physical sensory signals available on our hardware platforms.

\vspace{0.01in}
{\bf Implementation details.}
We train all models by fine-tuning GR00T N1.5 and $\pi_0$ on a consolidated dataset of training data from all tasks, following their official implementation: full fine-tuning for $\pi_0$ and training only the action expert for GR00T N1.5. For GR00T N1.5 + Tactile-VLA, we additionally unfreeze the VLM backbone, since Tactile-VLA injects tactile signals into the VLM and the backbone must adapt to this unseen modality.
We provide further implementation details in Appendix~\ref{appendix:impl_details}.

\subsection{Main Results}\label{sec:exp:main}
We provide quantitative results in \Cref{tab:main} and qualitative results in \Cref{fig:qualitative}. We also provide more results, including evaluation on simulated manipulation benchmarks in Appendix~\ref{appendix:sec:quantitative}.

In \Cref{tab:main}, we first observe that base VLAs (\ie fine-tuned VLAs without physical sensing) achieve low average success rates of 20.8\% and 27.1\% for GR00T N1.5 and $\pi_0$, respectively, confirming that vision and language alone are insufficient for tasks requiring fine-grained force control.
In contrast, we find that \sname{} achieves substantial improvements across all tasks and base models.
For example, for GR00T N1.5, we achieve an average improvement of 28.2\% over the base model, with particularly strong gains on Unstack Cup (+37.5\%) and Board Erase (+29.2\%).
Similar improvements are observed on $\pi_0$, demonstrating that our approach generalizes across different VLA architectures. 
We also find that \sname{} consistently outperforms several physical modality-augmented VLA baselines.
For example, on the PnP Egg task, GR00T N1.5 + \sname{} with tactile signals achieves 66.7\%, but GR00T N1.5 + ForceVLA gets 54.2\%.
This highlights our design that maintains a separate processing stream with controlled cross-modal joint self-attention, which enables effective integration of physical sensing without corrupting action representation.
Moreover, \sname{} shows consistent improvements when additional physical modalities are incorporated, demonstrating that \sname{} is a unified framework for adapting VLAs to multiple physical modalities.

\begin{figure}[t!]
    \centering\small
    \vspace{-1em}
    \includegraphics[width=0.95\linewidth]{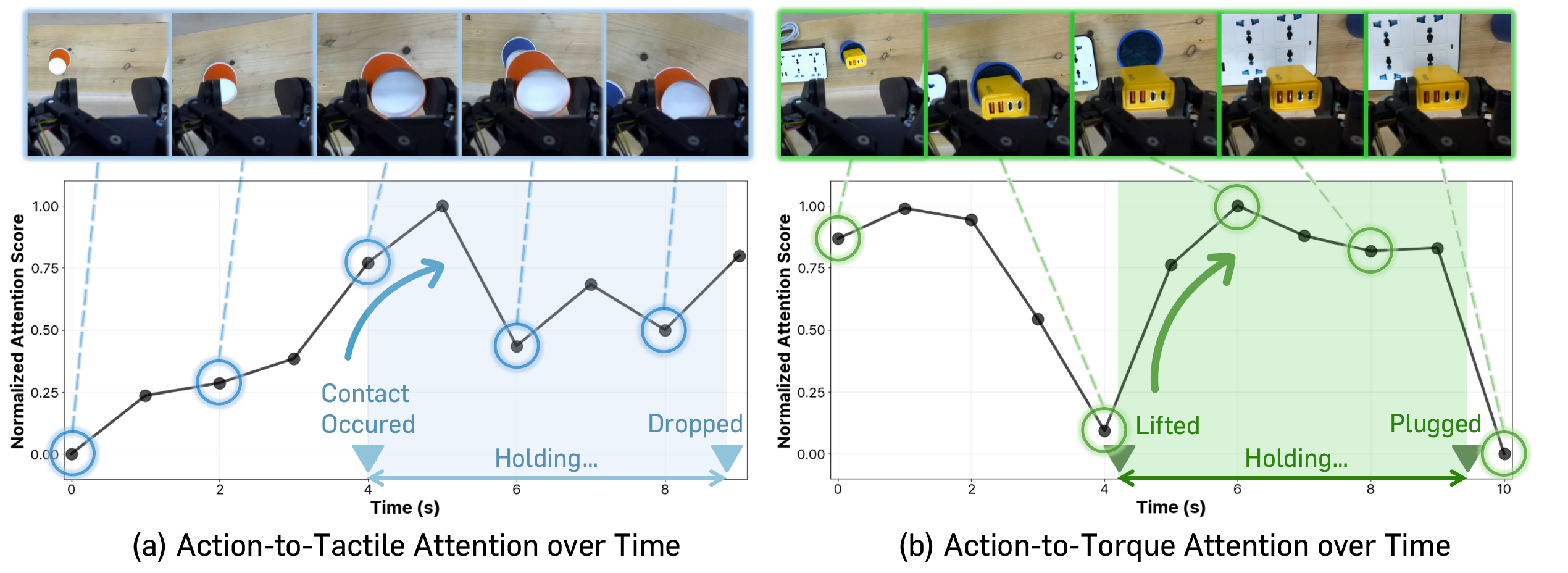}
    \caption{\textbf{Attention scores between action and physical modality stream.}
    We visualize the standardized attention scores between action and tactile/torque sensory streams in the joint cross-modal self-attention layers while conducting tasks. We find that action tokens attend strongly to tactile and torque tokens at moments that demand accurate, contact-intensive control, \eg (a) when tactile contact is detected, and (b) when the robot identifies the insertion point to plug in.
    }
    \label{fig:attention}
  % \vspace{-0.2in}
\end{figure}

\subsection{Analysis and Ablation Study}\label{sec:exp:analysis}
\vspace{0.01in}
{\bf When does the VLA attend to physical feedback?}
To investigate when the VLA attends to physical feedback, we  visualize the attention patterns in our cross-modal attention mechanism while performing the task. In \Cref{fig:attention}, we observe that action tokens attend strongly to torque and tactile tokens when the generated actions require accurate, contact-intensive control (\eg when grasping softly), enabling the model to modulate insertion or grasping force based on resistance feedback. Notably, this demonstrates that \sname{} truly leverages physical sensory signals when generating actions at moments that require accurate, contact-intensive control.

\vspace{0.01in}
{\bf Effect of decoupled multimodal stream architecture.}
We compare \sname{} against a na\"ive Diffusion Transformer (DiT) action expert that jointly predicts physical sensory signals within a single stream.
In \Cref{tab:ablation-stream}, we observe that our decoupled multimodal architecture significantly outperforms the DiT action expert.
For example, on the Unstack Cup task, it achieves 20.9\% improvements over the na\"ive DiT baseline.
These results indicate that decoupling distinct modalities to separate streams is crucial for effectively adapting VLAs to diverse physical sensory signals.

\vspace{0.01in}
{\bf Effect of two-stage training.}
We compare \sname{} trained with either single- or two-stage scheme, where for the single-stage training, we skip physical alignment stage and jointly fine-tunes the action and physical streams.
In \Cref{tab:ablation-training}, we find that a two-stage training scheme is critical for fine-tuning VLAs to adapt to new physical feedback modalities, \eg on the Unstack Cup task, it yields a 16.7\% improvement over single-stage training.

\vspace{0.01in}
{\bf Effect of future physical modality prediction.}
We quantify the effect of the future physical modality prediction objective in \Cref{tab:ablation-prediction}.
We compare \sname{} trained with and without the future prediction objective, and find this objective provides meaningful guidance for generating more accurate actions.
For instance, on the Unstack Cup task, the prediction objective improves performance by 8.4\% (45.8 $\rightarrow$ 54.2\%).
We further visualize \sname{}'s prediction dynamics and analyze how anticipated physical signals guide action generation in Appendix~\ref{appendix:sec:pred-dynamics}.

\begin{table*}[t]
  \centering
  \small
  \setlength{\tabcolsep}{4pt}
  \renewcommand{\arraystretch}{0.95}
  \vspace{-0.2em}
   \caption{\textbf{Ablation studies.} We report the success rates (\%) of various variants of \sname{} on contact-rich tasks using GR00T N1.5 as the pretrained VLA.
  Our default setting is highlighted in \colorbox{gray!10}{gray}.}
  \label{tab:ablation}
  \vspace{-0.4em}

  % ===== Row 1: (a) Stream | (b) Training | (c) Future prediction =====
  \begin{subtable}[t]{0.30\linewidth}
    \centering
    \caption{Stream design.}
    \vspace{-0.3em}
    \begin{tabular}{lcc}
      \toprule
      Stream & \makecell{Unstack\\Cup} & \makecell{PnP\\Egg} \\
      \midrule
      Single stream  & 33.3 & 50.0 \\
      \rowcolor{gray!10}
      Decoupled & \textbf{54.2} & \textbf{66.7} \\
      \bottomrule
    \end{tabular}
    \label{tab:ablation-stream}
  \end{subtable}
  \hfill
  \begin{subtable}[t]{0.30\linewidth}
    \centering
    \caption{Training scheme.}
    \vspace{-0.3em}
    \begin{tabular}{lcc}
      \toprule
      Training & \makecell{Unstack\\Cup} & \makecell{PnP\\Egg} \\
      \midrule
      One-stage  & 37.5 & 58.3 \\
      \rowcolor{gray!10}
      Two-stage & \textbf{54.2} & \textbf{66.7} \\
      \bottomrule
    \end{tabular}
    \label{tab:ablation-training}
  \end{subtable}
  \hfill
  \begin{subtable}[t]{0.31\linewidth}
    \centering
    \caption{Future prediction.}
    \vspace{-0.3em}
    \begin{tabular}{lcc}
      \toprule
      \makecell[l]{Future\\prediction} & \makecell{Unstack\\Cup} & \makecell{PnP\\Egg} \\
      \midrule
      w/o prediction & 45.8 & 58.3 \\
      \rowcolor{gray!10}
      w/ prediction  & \textbf{54.2} & \textbf{66.7} \\
      \bottomrule
    \end{tabular}
    \label{tab:ablation-prediction}
  \end{subtable}
\vspace{-0.1in}
\end{table*}
\vspace{0.01in}
{\bf Inference efficiency.}
\sname{} introduces additional processing for multiple physical sensory signals.
Then, how does this extension affect inference latency?
In \Cref{tab:efficiency}, we analyze the inference latency of GR00T N1.5 + \sname{} when incorporating multiple signals.
We find that when both tactile and torque modalities are incorporated, latency increases by only 2.4 ms (1.11$\times$ slower than the base model), highlighting the efficiency of \sname.
Importantly, \sname{} offers significant performance gain in real-world tasks while incurring only a marginal increase in inference latency.

\subsection{Applicability to Different Embodiment}\label{sec:exp:hand}

To demonstrate that \sname{} is not tied to a specific embodiment or physical feedback, we further evaluate GR00T N1.5 and GR00T N1.5 + MoSS on an OpenArm robot equipped with a 6-DoF Inspire RH56F1 dexterous hand. We use a 6-dimensional fingertip force signal provided by the hand. We design two contact-rich manipulation tasks on this setup (see \cref{fig:qualitative_hand} for the visualization of the tasks, Appendix~\ref{appendix:tasks} for more detailed descriptions, and implementation details in Appendix~\ref{appendix:impl_details}):
\vspace{-0.3em}
\begin{itemize}[leftmargin=*,itemsep=0mm]
    \item \textbf{Pick Paper.} A robot must pick up a thin sheet of paper (or napkin) from a flat surface. This task requires fine-grained force feedback at the fingertips to grasp the paper without slipping.

    \item \textbf{Sort by Weight.} A robot must lift a bag and place it right if heavy or left if light. This task requires sensing fingertip force during lifting to infer the bag's weight and decide where to place it.
\end{itemize}

\begin{figure}[t]
\begin{minipage}[t]{0.52\linewidth}
    \centering
    \captionof{table}{\textbf{Efficiency analysis.}
    We report the inference latency (ms) of GR00T N1.5 + \sname{} under different physical input configurations measured per action chunk, with relative overhead shown in parentheses.}
    \label{tab:efficiency}
    \vspace{-0.1em}
    \resizebox{0.95\linewidth}{!}{
      \begin{tabular}{lccc}
        \toprule
        Method & Tactile & Torque & Latency (ms, $\downarrow$) \\
        \midrule
        GR00T N1.5 & \xk & \xk & 21.0 (1.00$\times$) \\
        \arrayrulecolor{black!40}\midrule
        \multirow{3}{*}{\textbf{+ \sname{} (Ours)}} & \ck & \xk & 22.4 (1.06$\times$) \\
        & \xk & \ck & 21.9 (1.04$\times$) \\
        & \ck & \ck & 23.4 (1.11$\times$) \\
        \arrayrulecolor{black}
        \bottomrule
      \end{tabular}
    }
\end{minipage}
\hfill
\begin{minipage}[t]{0.46\linewidth}
    \captionof{table}{\textbf{Applicability to a dexterous hand embodiment.}
    We evaluate \sname{} on an OpenArm + Inspire RH56F1 Hands with fingertip force (6-dim) sensors, and report success rates (\%) over 24 trials per task.}
    \label{tab:hand}
    \vspace{-0.4em}
    \small
    \centering
    \resizebox{0.95\linewidth}{!}{
    \begin{tabular}{lccc}
        \toprule
        Method & \makecell{Pick\\Paper} & \makecell{Sort by\\Weight} & \textbf{Avg.} \\
        \midrule
        GR00T N1.5 & 62.5 & 54.2 & 58.4 \\
        \rowcolor{gray!10}
        + \textbf{\sname{} (Ours)} & \textbf{83.3} & \textbf{95.8} & \textbf{89.6} \\
        \bottomrule
    \end{tabular}}
\end{minipage}
\end{figure}
In \Cref{tab:hand}, we find that \sname{} consistently outperforms GR00T N1.5 across both tasks. In particular, we find a notable gain on Sort by Weight, where successful execution requires actively reasoning about the perceived force control. This indicates that MoSS effectively leverages physical feedback even on a substantially different embodiment.

\section{Conclusion}

In this work, we have presented \sname, a unified framework that adapts Vision-Language-Action models (VLAs) to leverage multiple sensory signals for action prediction. By extending the action experts with new streams, \sname{} can incorporate various modalities without corrupting the pretrained features.
Our experiments show that \sname{} improves performance on contact-rich, real-world manipulation tasks, and the performance continues to improve as more physical sensory modalities are incorporated.
We hope our approach facilitates future research toward generalist robot policies that can leverage a diverse range of physical sensory modalities.

\section{Limitations}

{\textbf{Practical considerations for sensor deployment.}
While our experiments confirm the effectiveness of incorporating tactile, torque, and force signals into \sname{}, physical sensors in general are known to exhibit wear or noise over extended use, which may need to be considered when deploying our framework in long-term, real-world scenarios. Pairing \sname{} with higher-fidelity, more durable sensors and periodic recalibration would help ensure consistent performance in such settings.}

\textbf{Dependence on the base policy for non-contact phases.}
\sname{} complements a pretrained VLA once meaningful physical interaction occurs, rather than enhancing VLAs' pre-contact visual perception and action generation. Accordingly, if the base policy fails to reach the object or form a stable initial grasp, \sname{} provides limited corrective information. Addressing these failures will likely require complementary advances in the base policy's visual perception and motion planning. We provide a more detailed discussion of failure cases in Appendix~\ref{appendix:sec:failure_case_analysis}.

\clearpage
% % The acknowledgments are automatically included only in the final and preprint versions of the paper.
% \acknowledgments{If a paper is accepted, the final camera-ready version will (and probably should) include acknowledgments. All acknowledgments go at the end of the paper, including thanks to reviewers who gave useful comments, to colleagues who contributed to the ideas, and to funding agencies and corporate sponsors that provided financial support.}

%===============================================================================

\bibliography{reference}  % .bib
\newpage
\appendix
\onecolumn

\section{Experimental Details}

\subsection{Task Specifications}\label{appendix:tasks}
We collect 100 demonstrations per task for Unstack Cup, PnP Egg, Board Erase, Plug Insertion, and Pick Paper, and 60 demonstrations for Sort by Weight, with extensive variation across multiple dimensions to ensure robust generalization and avoid memorization. For the single-arm gripper setup, we use a third-person view camera and a wrist-mounted camera, both at 720$\times$1280 resolution without depth. For the bimanual dexterous-hand setup, we use a stereo egocentric camera at the same resolution without depth. All visual observations are resized to 224$\times$224 before being fed into the model. We use absolute joint position as the action mode. 
We describe the detailed scenarios as follows:

\textbf{Unstack Cup:} The task required lifting only the top red cup from a stack. To prevent gripper position memorization, we varied the cup sizes, using both large and small cups in different configurations. In particular, we evaluate each model over 24 trials, consisting of 12 trials with a large cup and 12 trials with a small cup. We use the instruction ``\textit{Pick up the top red cup from the stack and place it next to the blue cup.}" for this task. 

\textbf{PnP Egg:} The egg was placed in various positions across the workspace to evaluate spatial generalization. We evaluate each model over 24 trials. We use the instruction ``\textit{Pick up the egg and place it in the gold bowl.}" for this task.

\textbf{Board Erase:} The whiteboard was positioned at three different heights to test height adaptability. We varied the eraser colors (red and grey) with slight differences in object size, and used markers of different colors (blue and black). Evaluations were conducted across diverse environmental conditions to assess robustness. In particular, we evaluate each model over 24 trials, consisting of one trial per combination of (height, eraser color, object size, different color).
We use the instruction ``\textit{Use the (red/grey) eraser to clean the whiteboard.}" for this task.

\textbf{Plug Insertion:} We used chargers of three different colors (yellow, black, and white) for color and object generalization, with target insertion positions varied across trials. We evaluate each model over 24 trials, consisting of 8 trials per charger. We use the instruction ``\textit{Pick up the (yellow/white/black) charger and plug it into the socket.}" for this task.

\textbf{Pick Paper:} The task is performed on a height-adjustable table, allowing us to test height adaptability by varying the table height across trials. Specifically, we evaluate with three different heights: one that is in-domain with respect to the training data, and two that are out-of-domain, both lower than any height seen during training. Since the policy cannot rely on proprioception alone to localize the surface at unseen heights, this setup requires the robot to use fingertip force feedback to detect contact with the paper (or napkin). Evaluations are further conducted across diverse paper positions on the table. We evaluate each model over 24 trials. We use the instruction ``\textit{Pick up the paper.}'', ``\textit{Pick up the napkin.}''  for this task.

\textbf{Sort by Weight:} A single camping bag is used across all trials, and its weight is varied by placing 0, 1, or 2 juice pouches (each approximately 200g) inside, yielding three weight conditions. The robot must lift the bag, infer its weight from fingertip force feedback during lifting, and place it on the right if heavy or on the left if light. We evaluate each model over 24 trials across the three weight conditions. We use the instruction ``\textit{Place the bag on the right if heavy, and on the left if light.}'' for this task.

\newpage

\begin{figure}[t]
    \centering\small
    \includegraphics[width=.97\linewidth]{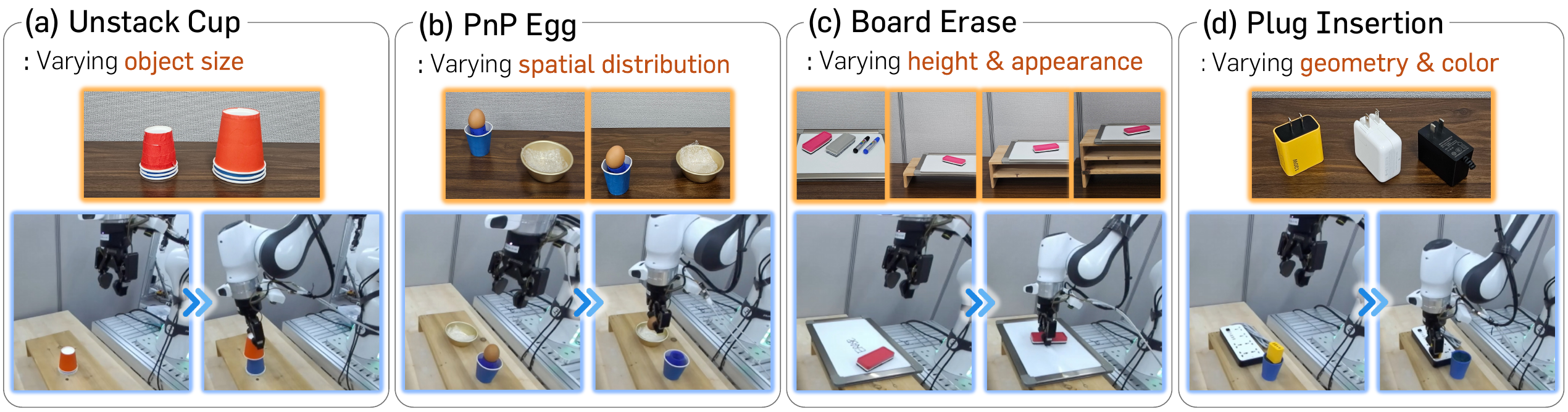}
    \vspace{-0.1cm}
    \caption{\textbf{Examples of the evaluation tasks.}
    We design contact-rich, real-world robotic manipulation tasks with systematically controlled variations in object size, spatial layout, geometry, and appearance, making visual observations alone ambiguous and physical feedback necessary for success.
    }
    \label{fig:task_overview}
\end{figure}
\begin{figure}[t]
    \centering\small
    \includegraphics[width=.99\linewidth]{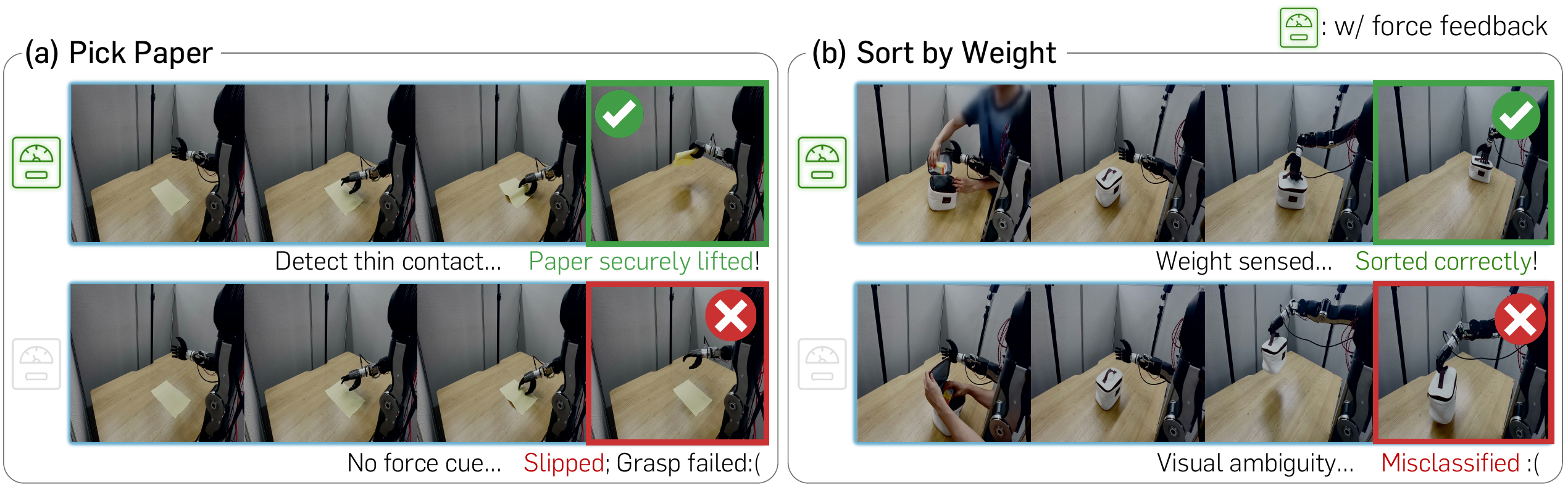}
    \caption{\textbf{Example rollouts of real-world dexterous hand tasks.}
    We provide example rollouts on hand tasks that critically depend on physical feedback, comparing \sname{} (top, with force feedback) against GR00T N1.5 (bottom, vision only).
    In (a) \textit{Pick Paper}, \sname{} detects slip and regulates grasp force to lift the flat paper, while the baseline fails to maintain a stable grasp.
    In (b) \textit{Sort by Weight}, \sname{} infers the hidden juice box from contact force and sorts the bag to the correct side, whereas the baseline cannot disambiguate weight and sorts incorrectly.
    }
    \label{fig:qualitative_hand}

\end{figure}

\subsection{Hardware Setup: Sensor Specifications}\label{appendix:hardware}
\begin{figure*}[h]
    \centering\small
    \begin{subfigure}[t]{0.48\linewidth}
        \centering
        \includegraphics[width=\linewidth]{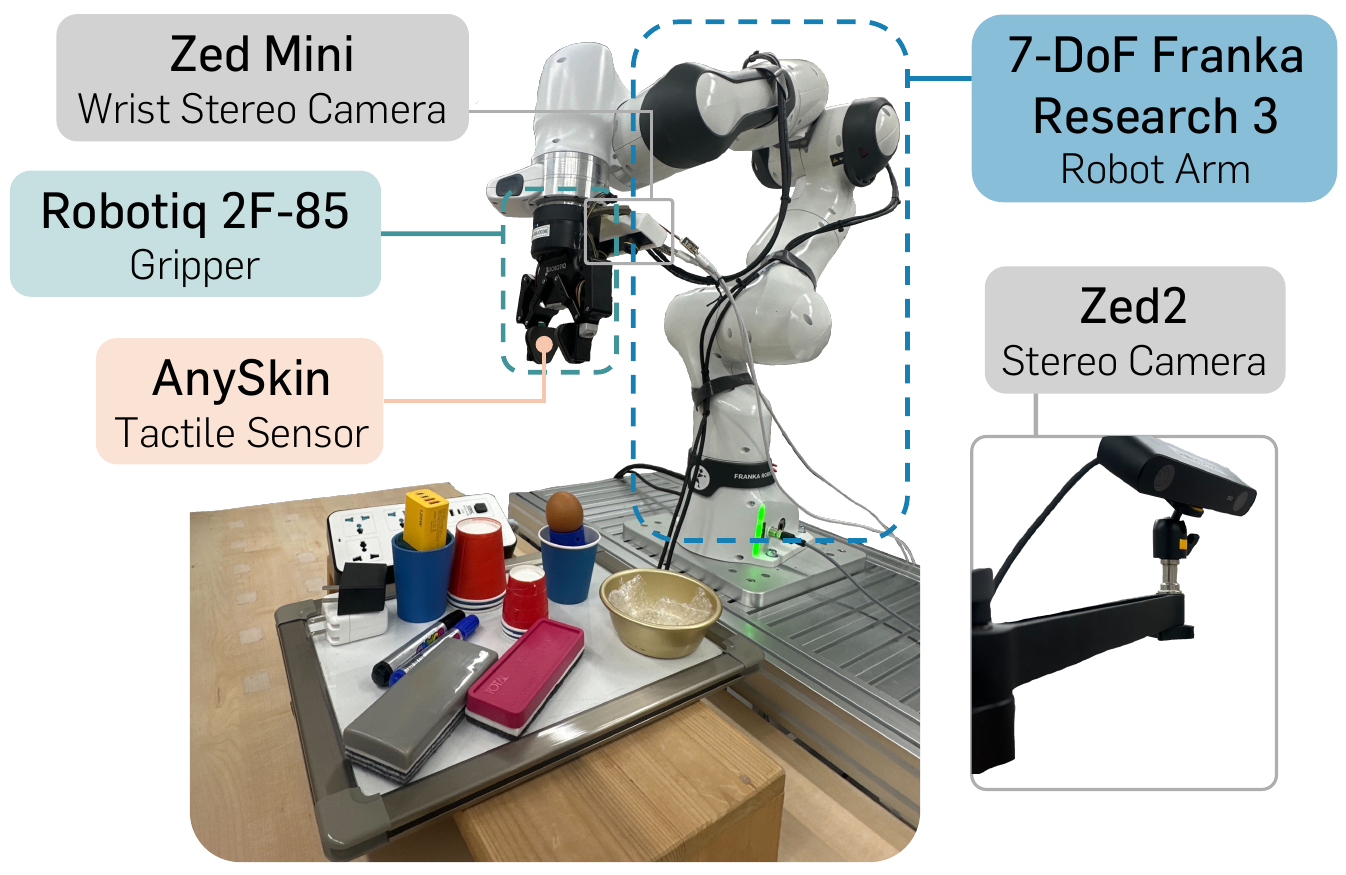}
        \caption{\textbf{Single-arm gripper platform.}
        A 7-DoF Franka Research 3 robot arm equipped with a Robotiq 2F-85 gripper and AnySkin tactile sensors mounted on the gripper fingers, together with a wrist-mounted Zed Mini stereo camera for visual observation.}
        \label{fig:hardware_franka}
    \end{subfigure}
    \hfill
    \begin{subfigure}[t]{0.48\linewidth}
        \centering
        \includegraphics[width=\linewidth]{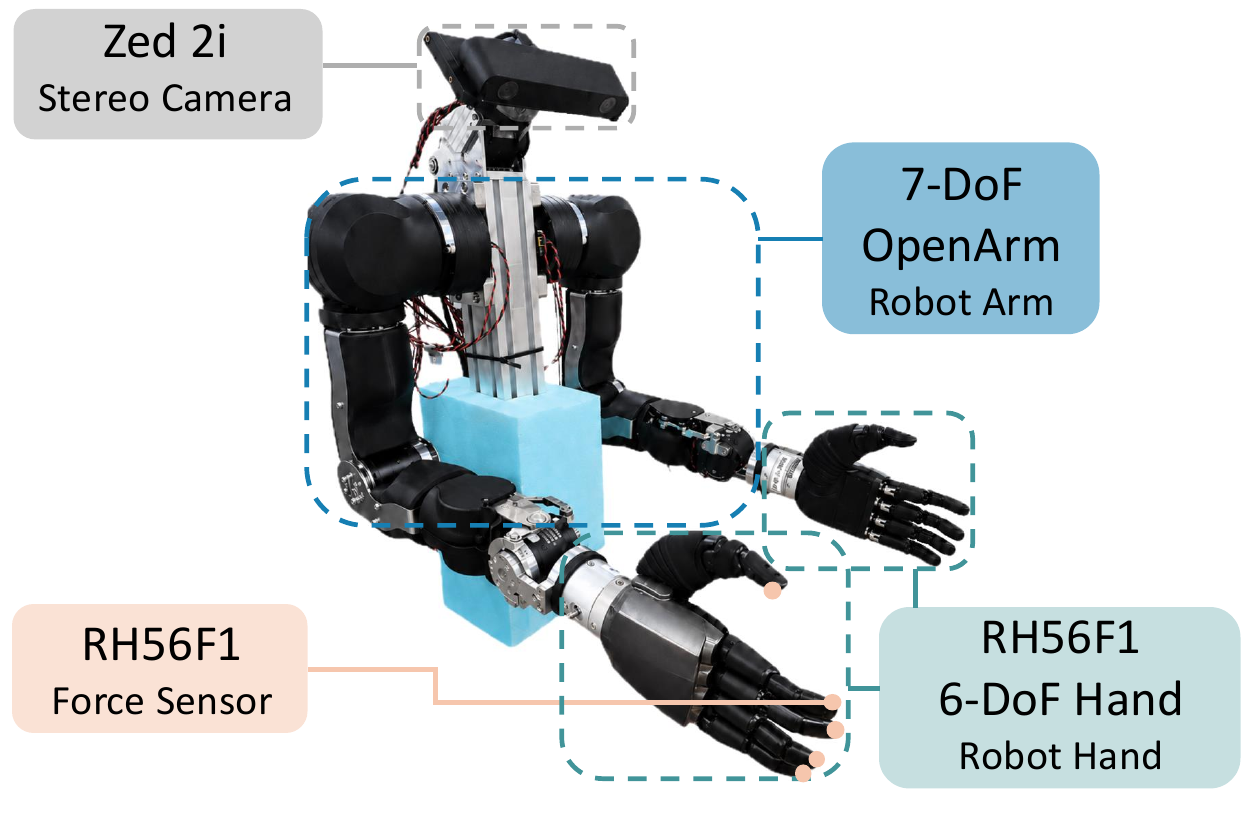}
        \caption{\textbf{Bimanual dexterous-hand platform.}
        Dual 7-DoF OpenArm robot arms, each equipped with a 6-DoF RH56F1 dexterous hand and an RH56F1 force sensor at each fingertips, together with a head-mounted Zed 2i stereo camera for visual observation.}
        \label{fig:hardware_openarm}
    \end{subfigure}
    \caption{\textbf{Overview of our hardware platforms.}
    We use two complementary robot setups that provide tactile, force, and torque sensing for contact-rich manipulation.}
    \label{fig:hardware}
\end{figure*}

We visualize the hardware setup of both platforms, including sensor specifications, in \Cref{fig:hardware}.
For the main experiments, we construct a Franka Research 3 platform, a 7-DoF single arm robot equipped with a Robotiq 2F-85 gripper. Here, we mount an AnySkin \citep{bhirangi2025anyskin} tactile sensor on the gripper, while the robot arm itself supports torque sensing at each joint.
Specifically, for tactile signals, each AnySkin sensor provides a 15-dimensional feature vector, obtained from five sensing units that each measure a 3-dimensional force vector (\ie $x$, $y$, $z$). Since we attach one AnySkin sensor to each gripper finger, our tactile sensing system provides a 30-dimensional observation vector at each timestep. For the torque signals, the robot provides a 7-dimensional vector (one scalar torque value per joint) measured in Newton-meters.
The dual-sensor configuration enables the system to simultaneously monitor forces applied by both gripper fingers, providing rich information about grasp stability and contact interactions with manipulated objects.
For the applicability to different embodiments, we consider the OpenArm robot equipped with an Inspire RH56F1 Hands, a 6-DoF dexterous hand with fingertip force sensors. The force sensor provides a 6-dimensional force signal (one per finger, with two on the thumb).

\newpage

\subsection{Implementation Details}\label{appendix:impl_details}

We report the performance of \sname{} by fine-tuning two pretrained Vision-Language-Action models, GR00T N1.5 and $\pi_0$, following their official implementations.
For GR00T N1.5, we freeze the VLM backbone and only train the action expert; for $\pi_0$, we fine-tune all model parameters.
The only exception is GR00T N1.5 + Tactile-VLA, where we additionally fine-tune the VLM backbone, since Tactile-VLA injects tactile signals directly into the VLM and the backbone must adapt to this previously unseen modality.

All models are trained jointly on a consolidated dataset combining the training data from all tasks, with a batch size of 32.
For GR00T N1.5, we train for 60k iterations, and for $\pi_0$, we train for 30k iterations.
We set the coefficient of the physical prediction loss to $\lambda_{\mathrm{phy}}=0.1$.
The remaining training hyperparameters are summarized in Table~\ref{tab:hyperparams}.

\begin{table}[h]
    \centering
    \caption{Hyperparameter details for training \sname{}.}
    \label{tab:hyperparams}
    \begin{adjustbox}{max width=\textwidth}
    \begin{tabular}{lcc}
    \toprule
     & GR00T N1.5 + \sname{} & $\pi_0$ + \sname{} \\
    \midrule
    Optimizer & AdamW & AdamW \\
    Optimizer momentum & $\beta_1, \beta_2 = 0.95, 0.999$ & $\beta_1, \beta_2 = 0.9, 0.95$ \\
    Optimizer weight decay & 1e-5 & 1e-10 \\
    Learning rate & 1e-4 & 2.5e-5 \\
    Learning rate scheduler & Cosine decay & Cosine decay \\
    Warmup iterations & 3{,}000 & 1{,}000 \\
    Batch size & 32 & 32 \\
    Training iterations (stage 1) & 20{,}000 & 10{,}000 \\
    Training iterations (stage 2) & 40{,}000 & 20{,}000 \\
    \bottomrule
    \end{tabular}
    \end{adjustbox}
\end{table}

\newpage
\section{Additional Analysis}
\subsection{Hyperparameter Sensitivity Analysis }
\label{appendix:sec:loss_weight_ablation}
In \Cref{tab:ablation-loss-weight}, we analyze the hyperparameter sensitivity of training loss coefficients for the modular sensory streams $\lambda_{\text{phy}}$ on Unstack Cup and PnP Egg tasks. We observe that performance is relatively robust across the range $\lambda_{\text{phy}} \in \{0.1, 0.5\}$, with peak performance at $\lambda_{\text{phy}} = 0.1$. We also find that performance degrades at $\lambda_{\text{phy}} = 1.0$, where the future prediction loss increasingly dominates the action learning signals. Our default setting is highlighted in \colorbox{gray!10}{gray}.

  \begin{table}[h]
  \vspace{-0.2in}
    \centering
    \caption{Results of GR00T N1.5 + MoSS for different $\lambda_{\mathrm{phy}}$ values.}
    \begin{tabular}{lcc}
      \toprule
      $\lambda_{\mathrm{phy}}$ & \makecell{Unstack Cup} & \makecell{PnP Egg} \\
      \midrule
      \rowcolor{gray!10}
      0.1 & \textbf{54.2} & \textbf{66.7} \\
      0.5 & 50.0 & 58.3 \\
      1.0 & 29.2  & 41.7 \\
      \bottomrule
    \end{tabular}
    \label{tab:ablation-loss-weight}
  \end{table}

\subsection{Prediction Dynamics Analysis}
\label{appendix:sec:pred-dynamics}
\begin{figure*}[h]
    \centering\small
    \includegraphics[width=0.98\linewidth]{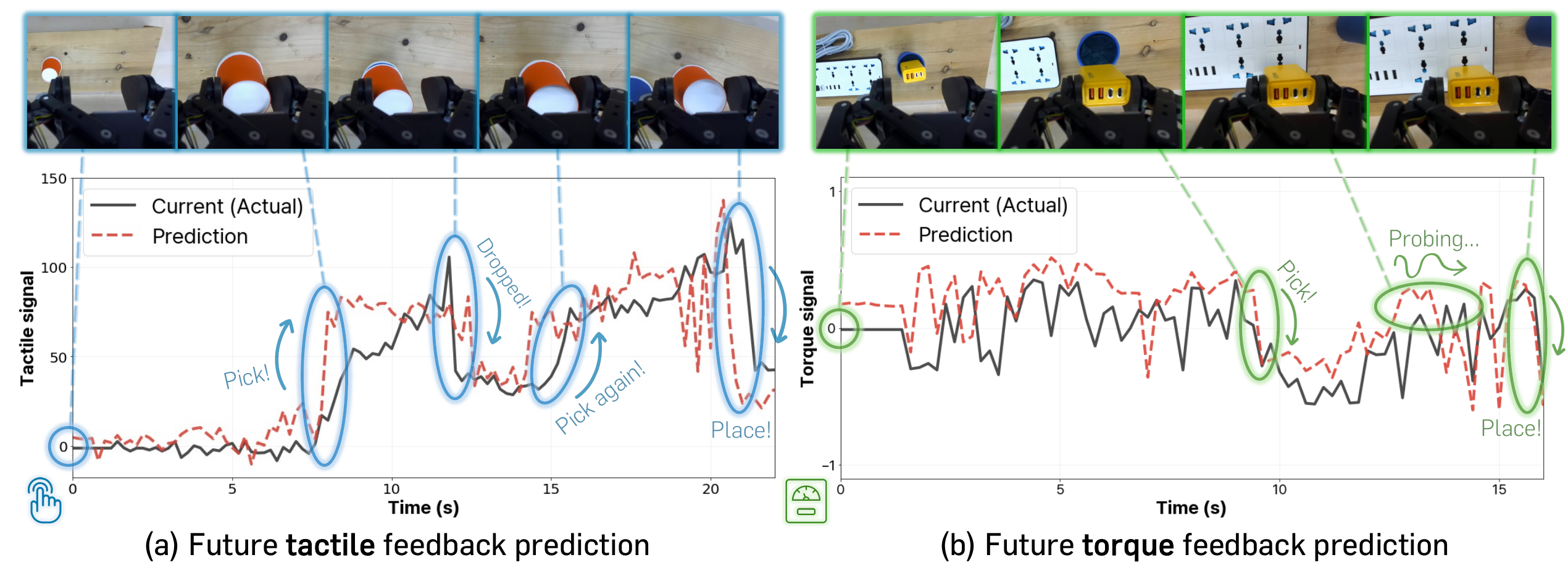}
    \caption{\textbf{Physical feedback and predicted values.}
    We visualize the physical sensory signals (\ie tactile and torque) predicted by \sname{} while conducting tasks.
    We find that \sname{} accurately predicts the changes in physical signals, particularly at moments when such signals are critical for successfully performing the task.
    }
    \label{fig:graph}
\end{figure*}

We visualize \sname{}'s prediction dynamics in \Cref{fig:graph}.
We find that \sname{} anticipates contact by predicting rising tactile and torque signals, prompting increased gripping force.
This demonstrates how the future prediction objective provides grounded physical guidance that leads to the performance gains reported in \Cref{tab:ablation-prediction}.

\newpage

\subsection{Failure Case Analysis}
\label{appendix:sec:failure_case_analysis}
While MoSS demonstrates meaningful improvement on contact-rich tasks, we find several failure cases, which are visualized in  \Cref{fig:failure_case_examples}. First, in tasks such as Unstack Cup, PnP Egg, and Board Erase, we observe that when the base policy fails to reach the target object or establish a stable grasp, MoSS cannot compensate. This is because the physical feedback never becomes informative at these stages, since the gripper has not yet made meaningful contact. Moreover, in the Plug Insertion task, we observe that when the initial grasp is imperfect, it results in a misaligned plug angle that prevents successful socket insertion. As with the first case, this is also related to the grasping quality issue that occurs before physical feedback becomes effective.

We observe similar failures in the dexterous hand tasks. In the Pick Paper task, especially at lower table heights, the hand occasionally fails to apply sufficient normal force against the paper before picking up. This causes the paper to slip from the fingertips.
In the Sort by Weight task, the policy must first insert the hand into the bag handle and lift the bag to obtain informative fingertip force feedback for weight estimation. However, when the hand fails to grasp the handle properly, it cannot lift the bag and thus never receives the physical feedback required for the downstream decision.
These examples highlight that \sname{} is most effective after meaningful contact is established, but remains dependent on the base policy for pre-contact grasping and alignment.

\begin{center}
  \includegraphics[width=\linewidth]{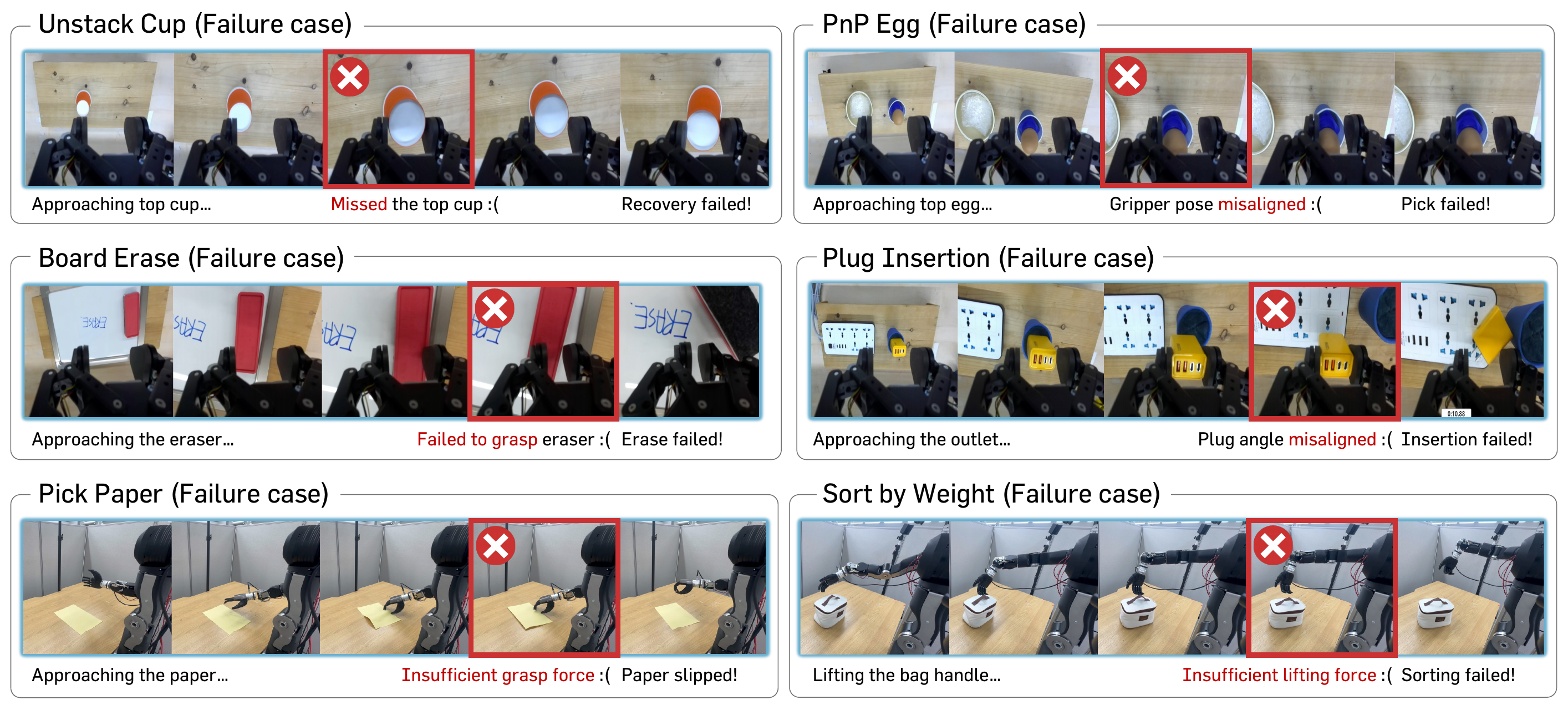}
  \captionof{figure}{\textbf{Failure case examples.}}
  \label{fig:failure_case_examples}
\end{center}

\newpage

\section{More Discussion on Related Work} \label{appendix:related-work}

\vspace{0.01in}
{\bf Multimodal architectures.}
Many recent efforts have attempted to design an architecture to incorporate a diverse range of modalities. In particular, the success of VLMs has been largely driven by advances in multimodal fusion architectures: early approaches~\citep{nagrani2021attention, wei2020multi} employed cross-attention mechanisms to align visual and textual representations, while more recent work has explored richer fusion strategies, including perceiver-based architectures~\citep{jaegle2021perceiver}, modality-specific adapters~\citep{li2023blip, yang2024mma}, and unified tokenization schemes~\citep{lu2022unified, zhou2024transfusion, liang2024mixture}.
These methods demonstrate that carefully designed fusion mechanisms can effectively model heterogeneous modalities with different dimensionalities and semantic structures.
In the robotics domain, recent VLAs~\citep{black2024pi_0, bjorck2025gr00t} have successfully adapted such vision-language fusion techniques to incorporate action prediction.
However, extending these architectures to physical sensing modalities poses additional challenges, as such signals differ fundamentally from visual and linguistic inputs in temporal structure, scale, and semantics.
Our work draws inspiration from multimodal fusion in VLMs, but adapts it specifically to the action prediction setting, introducing a modular architecture that enables pretrained VLAs to integrate physical sensory signals while preserving their action priors.

\vspace{0.01in}
{\bf Behavior cloning (BC) policy models with physical feedback.} 
Prior work augments behavior cloning and diffusion-based policies~\citep{chi2023diffusion} with physical feedback by either expanding the observation space with tactile or force inputs~\citep{xue2025reactive} or adding modality-specific encoders to the policy architecture~\citep{helmut2025tactile}, training the resulting multimodal policy from scratch.
These approaches typically rely on a single sensing modality at a time, such as a vision-based tactile sensor (\eg GelSight, DIGIT), or a skin-based tactile array~\citep{ablett2024multimodal, pattabiraman2024learning, zhao2025touch, xue2025reactive}, or a wrist- or fingertip-mounted force/torque sensor for contact-rich tasks such as peeling and insertion~\citep{liu2025forcemimic}.
While these works demonstrate the value of physical feedback for contact-rich manipulation, they are developed as task-specific policies tied to a fixed sensor configuration, without leveraging the broad semantic and behavioral priors of pretrained VLAs.
In contrast, our framework builds on pretrained VLAs and unifies heterogeneous physical feedback signals within a single architecture, enabling generalist policies that flexibly leverage available sensing modalities across various tasks and hardware configurations.

\newpage

\section{Additional Quantitative Results}
\label{appendix:sec:quantitative}

\subsection{Results on Simulated Robotic Manipulation Benchmarks}
In \Cref{tab:appendix_rlbench}, we report the performance of VLAs on the RLBench benchmark \citep{james2020rlbench}, across 9 tasks, using the force signals natively supported by the simulator. Among these, 4 tasks involve contact-sensitive interactions: open window (objects occluded during manipulation), move hanger (requiring delicate contact), and lamp on / push button (requiring precise pressing). We observe that MoSS significantly improves performance on the 4 contact-sensitive tasks, by improving GR00T N1.5 by 10.9\%. We also observe that MoSS maintains comparable performance on non-contact sensitive tasks. For instance, GR00T N1.5 achieves 46.8\% on the non-contact-sensitive tasks, while GR00T N1.5 + MoSS achieves 47.1\%. This demonstrates that MoSS effectively leverages physical signals for contact-sensitive tasks without degrading general manipulation capabilities.

\begin{table*}[h]
  \centering
  \caption{\textbf{Results on RLBench benchmark.}
  We report the success rates (\%) for each task using VLAs finetuned on a consolidated dataset that combines the training data from all tasks. We use the force signals natively supported by the simulator.
  Avg. denotes averaged success rates over entire tasks.
  \textbf{Bold} indicates best results.}
  \label{tab:appendix_rlbench}
  \begin{adjustbox}{max width=\textwidth}
  \begin{tabular}{lcccccccccccc}
    \toprule
    & \multicolumn{5}{c}{Contact-sensitive Tasks} & \multicolumn{6}{c}{Non-contact-sensitive Tasks} \\
    \cmidrule(lr){2-6} \cmidrule(lr){7-12}
    \shortstack{\\Method} & \shortstack{Open\\Window} & \shortstack{Lamp\\On} & \shortstack{Move\\Hanger} & \shortstack{Push\\Button} & \shortstack{\\\textbf{Avg.}} & \shortstack{Basketball\\in Loop} & \shortstack{Phone\\on Base} & \shortstack{Pick Up\\Cup} & \shortstack{Slide Block \\to Target} & \shortstack{Take Lid Off\\ Saucepan} & \shortstack{\\\textbf{Avg.}} & \shortstack{\textbf{Overall}\\\textbf{Avg.}} \\
    \midrule
    GR00T N1.5 \citep{nvidia2025gr00t} & 8.0 & 36.0 & 29.3 & 36.0 & 27.3 & \textbf{72.0} & 20.7 & 50.0 & 18.7 & \textbf{72.7} & 46.8 & 38.1 \\
    \cellcolor{gray!10}\textbf{+ MoSS (Ours)} & \cellcolor{gray!10}\textbf{9.3} & \cellcolor{gray!10}\textbf{47.3} & \cellcolor{gray!10}\textbf{45.3} & \cellcolor{gray!10}\textbf{50.7} & \cellcolor{gray!10}\textbf{38.2} & \cellcolor{gray!10}70.0 & \cellcolor{gray!10}\textbf{22.7} & \cellcolor{gray!10}\textbf{52.7} & \cellcolor{gray!10}\textbf{21.3} & \cellcolor{gray!10}68.7 & \cellcolor{gray!10}\textbf{47.1} & \cellcolor{gray!10}\textbf{43.1} \\
    \bottomrule
  \end{tabular}
  \end{adjustbox}
\end{table*}

\subsection{Scalability of VLAs to Multiple Physical Sensory Signals}
In \Cref{tab:multiple}, we report the performance of VLAs fine-tuned to incorporate multiple physical sensory signals. We find that, while existing approaches often degrade when additional modalities are introduced, \sname{} consistently improves as each new modality is added.

\begin{table*}[h]
  \centering
  \caption{\textbf{Results on using both tactile and torque inputs.}
    We report the success rates (\%) for each single-arm, gripper-based task when \emph{both} tactile and torque signals are provided as inputs to all methods.
    Values in parentheses indicate the absolute performance change compared to the corresponding single-modality setting in \cref{tab:main}.
    \textbf{Bold} and \underline{underline} indicate best and runner-up results, respectively.}
  \label{tab:multiple}
  \begin{adjustbox}{max width=\textwidth}
  \begin{tabular}{lccccccc}
    \toprule
    Method
      & Tactile & Torque
      & Unstack Cup & PnP Egg & Board Erase
      & Plug Insertion & Avg. \\
    \midrule
    GR00T N1.5~\citep{nvidia2025gr00t}
      & \xk & \xk
      & 16.7\phantom{($+$0.0)} & 45.8\phantom{($+$00.0)} & 20.8\phantom{($+$00.0)} & \phantom{0}0.0\phantom{($+$0.0)}
      & 20.8\phantom{($+$00.0)} \\

    \arrayrulecolor{black!40}\midrule
    \multirow{2}{*}{+ Tactile-VLA~\citep{huang2025tactile}}
      & \ck & \xk
      & 29.2\phantom{($+$0.0)} & 45.8\phantom{($+$00.0)} & 33.3\phantom{($+$00.0)} & 12.5\phantom{($+$0.0)}
      & 30.2\phantom{($+$00.0)} \\
      & \cellcolor{gray!10}\ck & \cellcolor{gray!10}\ck
      & \cellcolor{gray!10}12.5 \textcolor{BrickRed}{($-$16.7)} & \cellcolor{gray!10}37.5\phantom{0} \textcolor{BrickRed}{($-$8.3)} & \cellcolor{gray!10}16.7 \textcolor{BrickRed}{($-$16.6)} & \cellcolor{gray!10}16.7 \textcolor{blue}{($+$4.2)} & \cellcolor{gray!10}20.9\phantom{0} \textcolor{BrickRed}{($-$9.3)} \\
    \midrule
    \multirow{2}{*}{+ ForceVLA~\citep{yu2025forcevla}}
      & \ck & \xk
      & 37.5\phantom{($+$00.0)} & 54.2\phantom{($+$00.0)} & 33.3\phantom{($+$00.0)} & 12.5\phantom{($+$0.0)}
      & 34.4\phantom{($+$00.0)} \\
      & \cellcolor{gray!10}\ck & \cellcolor{gray!10}\ck
      & \cellcolor{gray!10}25.0 \textcolor{BrickRed}{($-$12.5)} & \cellcolor{gray!10}33.3 \textcolor{BrickRed}{($-$20.9)} & \cellcolor{gray!10}20.8 \textcolor{BrickRed}{($-$12.5)} & \phantom{0}\cellcolor{gray!10}4.2 \textcolor{BrickRed}{($-$8.3)} & \cellcolor{gray!10}21.9 \textcolor{BrickRed}{($-$12.5)} \\
    \midrule

    \multirow{2}{*}{+ TA-VLA~\citep{zhang2025ta}}
      & \xk & \ck
      & 29.2\phantom{($+$00.0)} & 45.8\phantom{($+$00.0)} & 37.5\phantom{($+$00.0)} & 20.8\phantom{($+$0.0)}
      & 33.3\phantom{($+$00.0)} \\
      & \cellcolor{gray!10}\ck & \cellcolor{gray!10}\ck
      & \cellcolor{gray!10}20.8\phantom{0} \textcolor{BrickRed}{($-$8.4)} & \cellcolor{gray!10}41.7\phantom{0} \textcolor{BrickRed}{($-$4.1)} & \cellcolor{gray!10}29.2\phantom{0} \textcolor{BrickRed}{($-$8.3)} & \cellcolor{gray!10}16.7 \textcolor{BrickRed}{($-$4.1)} & \cellcolor{gray!10}27.1\phantom{0} \textcolor{BrickRed}{($-$6.2)} \\
    \midrule
    \multirow{3}{*}{\textbf{+ \sname{} (ours)}}
      & \ck & \xk
      & \underline{45.8}\phantom{($+$00.0)} & \underline{66.7}\phantom{($+$00.0)} & \underline{41.7}\phantom{($+$00.0)} & 16.7\phantom{($+$0.0)}
      & \underline{42.7}\phantom{($+$00.0)} \\
      & \xk & \ck
      & 33.3\phantom{($+$00.0)} & 50.0\phantom{($+$00.0)} & \underline{41.7}\phantom{($+$00.0)} & \underline{25.0}\phantom{($+$0.0)}
      & 37.5\phantom{($+$00.0)} \\
      & \cellcolor{gray!10}\ck & \cellcolor{gray!10}\ck
      & \cellcolor{gray!10}\textbf{54.2\phantom{0} \textcolor{blue}{($+$8.4)}} & \cellcolor{gray!10}\textbf{66.7\phantom{0} \textcolor{black}{($-$0.0)}} & \cellcolor{gray!10}\textbf{50.0\phantom{0} \textcolor{blue}{($+$8.3)}} & \cellcolor{gray!10}\textbf{25.0 \textcolor{black}{($-$0.0)}} & \cellcolor{gray!10}\textbf{49.0\phantom{0} \textcolor{blue}{($+$6.3)}} \\
    \arrayrulecolor{black}\bottomrule
  \end{tabular}
  \end{adjustbox}
\end{table*}

\newpage 

\subsection{Full results}
In \Cref{tab:appendix_full}, we report the full performance of all VLAs, including the results of each fine-tuned model on several variations.

\begin{table*}[h]
  \centering
  \caption{\textbf{Full results on contact-rich, real-world manipulation tasks.}
  We report the success rates (\%) for each task using VLAs fine-tuned on a consolidated dataset that combines the training data from all tasks.
  For the tactile modality, we mount an AnySkin \citep{bhirangi2025anyskin} tactile sensor on the gripper, and for the torque modality, we use the joint torque measurements provided by the robot.
  Avg. denotes averaged success rates over entire tasks.
  \textbf{Bold} indicates best results.}
  \label{tab:appendix_full}
  \begin{adjustbox}{max width=\textwidth}
  \begin{tabular}{lcccccccccccc}
    \toprule
      &  & 
      & \multicolumn{2}{c}{\textbf{Unstack Cup}} & & \multicolumn{3}{c}{\textbf{Board Erase}}
      & \multicolumn{3}{c}{\textbf{Plug Insertion}} \\
      \cmidrule(lr){4-5} \cmidrule(lr){7-9} \cmidrule(lr){10-12}
    Method & Tactile & Torque & 
    Small & Big & \textbf{PnP Egg} &
    Low & Middle & High &
    Yellow & White & Black & \textbf{Avg.} \\
    \midrule
        GR00T N1.5~\citep{nvidia2025gr00t}
      & \xk & \xk
      &  \phantom{0}0.0 & 33.3 & 45.8 & \phantom{0}0.0 & 12.5 & \bf{50.0} & \phantom{0}0.0 & \phantom{0}0.0 & \phantom{0}0.0 & 20.8 \\
    \arrayrulecolor{black!40}\midrule
    + Tactile-VLA~\citep{huang2025tactile}
      & \ck & \xk
      & \phantom{0}8.3 & 50.0 & 45.8 & 12.5 & 37.5 & \bf{50.0} & 25.0 & 12.5 & \phantom{0}0.0 & 30.2   \\ 
    + ForceVLA~\citep{yu2025forcevla}
      & \ck & \xk
      & 16.7 & 58.3 & 54.2 & 12.5 & 37.5 & \bf{50.0} & 12.5 & 25.0 & \phantom{0}0.0 & 34.4 \\
    \textbf{+ \sname{} (ours)}
      & \ck & \xk
      & \bf{25.0} & 66.7 & \bf{66.7} & \bf{37.5} & 50.0 & 37.5 & 25.0 & 12.5 & \bf{12.5} & 42.7 \\ 
    \midrule
    + TA-VLA~\citep{zhang2025ta}
      & \xk & \ck
      & \phantom{0}8.3 & 50.0 & 45.8 & \bf{37.5} & 50.0 & 25.0 & 25.0 & 25.0 & \bf{12.5} & 33.3\\
    \textbf{+ \sname{} (ours)}
      & \xk & \ck
      & \phantom{0}8.3 & 58.3 & 50.0 & \bf{37.5} & \bf{62.5} & 25.0 & 25.0 & \bf{37.5} & \bf{12.5} & 37.5\\
    \midrule
    \rowcolor{gray!10}
    \textbf{+ \sname{} (ours)}
      & \ck & \ck
      & \bf{25.0} & \bf{83.3} & \bf{66.7} & \bf{37.5} & \bf{62.5} & \bf{50.0} & \bf{37.5} & 25.0 & \bf{12.5} & \bf{49.0} \\    

     \arrayrulecolor{black}\midrule
    
    $\pi_{0}$~\citep{black2024pi_0}
      & \xk & \xk
      & \phantom{0}0.0 & 25.0 & 50.0 & 12.5 & 37.5 & 37.5 & 25.0 & 12.5 & 12.5 & 27.1\\
    \arrayrulecolor{black!40}\midrule
     + Tactile-VLA~\citep{huang2025tactile}
      & \ck & \xk
      & \phantom{0}0.0 & 33.3 & 50.0 & 25.0 & 37.5 & 50.0 & 25.0 & 12.5 & \phantom{0}0.0 &  29.2\\
     + ForceVLA~\citep{yu2025forcevla}
      & \ck & \xk
      & \bf{16.7} & 33.3 & 62.5 & 50.0 & \bf{50.0} & 50.0 & 25.0 & 12.5 & 12.5 & 38.6\\
     \textbf{+ \sname{} (ours)}
      & \ck & \xk
      & \phantom{0}8.3 & \bf{50.0} & 62.5 & 25.0 & \bf{50.0} & 50.0 & 25.0 & 25.0 & 12.5 & 38.6\\
     \midrule
     + TA-VLA~\citep{zhang2025ta}
      & \xk & \ck
      & \phantom{0}8.3 & 25.0 & 58.3 & 12.5 & 37.5 & \bf{75.0} & 37.5 & 12.5 & 12.5 & 34.4\\
     \textbf{+ \sname{} (ours)}
      & \xk & \ck
      & \phantom{0}0.0 & 41.7& 62.5 & 50.0 & \bf{50.0} & 62.5 & \bf{50.0} & \bf{25.0} & 12.5 & 41.7\\
     \midrule
     \rowcolor{gray!10}
     \textbf{+ \sname{} (ours)}
      & \ck & \ck
      & \phantom{0}8.3 & \bf{50.0} & \bf{66.7} & \bf{62.5} & \bf{50.0} & 62.5 & 37.5 & \bf{25.0} & \bf{25.0} & \bf{45.9}\\

    \arrayrulecolor{black}\bottomrule
  \end{tabular}
  \end{adjustbox}
\end{table*}

\end{document}